%% file: bmvc_review.tex
\documentclass{bmvc2k}
\input{packages.tex}
\title{The Good, the Bad and the Ugly: Evaluating Convolutional Neural Networks for Prohibited Item Detection Using Real and Synthetically Composited X-ray Imagery}

\addauthor{Neelanjan Bhowmik$^1$}{}{1}
\addauthor{Qian Wang$^1$}{}{1}
\addauthor{Yona Falinie A. Gaus$^1$}{}{1}
\addauthor{Marcin Szarek$^3$}{}{3}
\addauthor{Toby P. Breckon$^{12}$}{}{12}

\addinstitution{
 Department of \\ 
 \{Computer Science$^1$ $|$ Engineering$^2$\} \\
 Durham University\\
 Durham, UK \\ 
 \vspace{0.2cm}
 $^3$School of Engineering \\ 
 Cranfield University\\
 Cranfield, UK 
 }
\runninghead{Bhowmik et al.}{X-Ray Prohibited Item Detection}

\def\etal{\emph{et al}\bmvaOneDot}
\begin{document}

\maketitle
\input{abstract.tex}
\input{1.introduction.tex}
\input{2.relatedwork.tex}

\input{3.proposal.tex}
\input{4.experimen.tex}
\input{5.evaluation.tex}
\input{6.conclusion.tex}
\input{7.reference.tex}

\end{document}

%% file: packages.tex
\usepackage{times}
\usepackage{epsfig}
\usepackage{graphicx}
\usepackage{amsmath}
\usepackage{amssymb}
\usepackage{multirow}
\usepackage{placeins}
\usepackage{wrapfig}
\usepackage[export]{adjustbox}
\usepackage{lettrine}
\usepackage{todonotes}

%% file: abstract.tex
\begin{abstract}
Detecting prohibited items in X-ray security imagery is pivotal in maintaining border and transport security against a wide range of threat profiles. Convolutional Neural Networks (CNN) with the support of a significant volume of data have brought advancement in such automated prohibited object detection and classification. However, collating such large volumes of X-ray security imagery remains a significant challenge. This work opens up the possibility of using synthetically composed imagery, avoiding the need to collate such large volumes of hand-annotated real-world imagery. Here we investigate the difference in detection performance achieved using real and synthetic X-ray training imagery for CNN architecture detecting three exemplar prohibited items, \big\{{\it Firearm, Firearm Parts, Knives}\big\}, within cluttered and complex X-ray security baggage imagery. 
We achieve 0.88 of mean average precision (mAP) with a Faster R-CNN and ResNet$_{101}$ CNN architecture for this 3-class object detection using real X-ray imagery. While the performance is comparable with synthetically composited X-ray imagery (0.78 mAP), our extended evaluation demonstrates both challenge and promise of using synthetically composed images to diversify the X-ray security training imagery for automated detection algorithm training.

\end{abstract}


%% file: 1.introduction.tex
\section{Introduction} \label{sec:intro}
\begin{figure}[h]
\includegraphics[width=\linewidth]{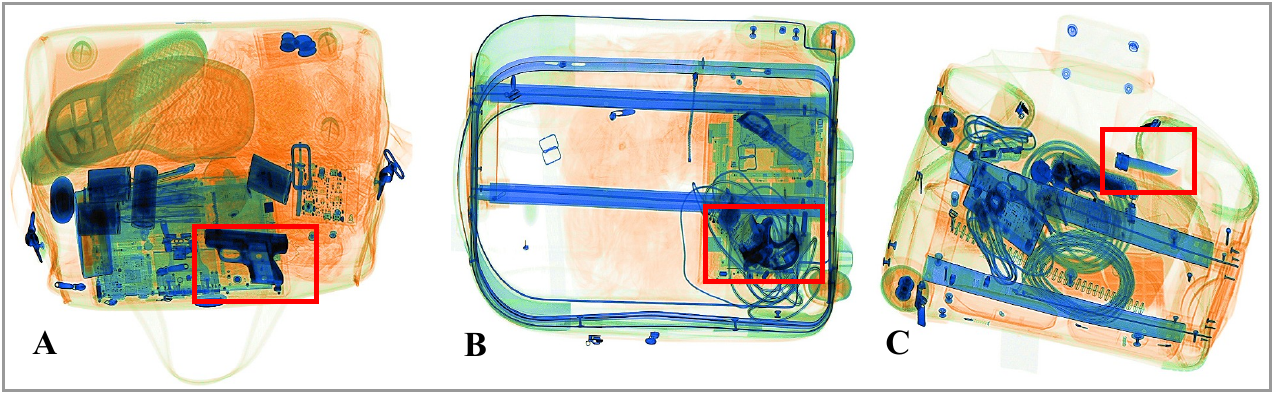}
 \vspace{-0.8cm}
 \caption{Exemplar X-ray security baggage images with prohibited objects - red box: (A) Firearm (B) Firearm Parts and (C) Knife.}
 \vspace{-0.3cm}
 \label{Fig:SOC_approach}
\end{figure}

To ensure transport and border security, X-ray security screening is commonplace within public transport and border security installations such as airports, railway and metro stations. However, due to the nature of cluttered and complex X-ray imagery (Figure \ref{Fig:SOC_approach}), the process of X-ray screening is complicated by tightly packed items within baggage making it challenging and time-consuming to identify the presence of prohibited items. With the natural occurrence of such prohibited items being rare, previous studies cite time constraints as a major factor in the performance limitations of human operators for this screening task \cite{Schwaninger2008Impact, Blalock2007Impact}.


Whilst challenging for a human, a reliable automatic prohibited item detection system may assist in improving the performance and throughput of such screening processes \cite{Turcsany2013Improving}. To date, contemporary X-ray security scanners already implement material discrimination via dual-energy multiple view X-ray imagery to enable threat material detection \cite{mouton2015review}. This use of dual-energy X-ray gives rise to the false-colour mapped appearance of X-ray security imagery  (e.g., metals, alloy or hard plastic are shown in blue while less dense objects are shown in green/orange - see Figure \ref{Fig:SOC_approach}).

Convolutional Neural Network (CNN) based methods have proven effective in detecting a wide range of object classes within this context  \cite{krizhevsky2012imagenet, he2016deep, szegedy2017inception,ren2015faster}. However, the performance of such object detection approaches is heavily reliant on the availability of a substantial volume of labelled X-ray imagery. Unfortunately, the availability of such X-ray imagery datasets suitable for training CNN architectures is limited and also restricted in size and item coverage (e.g. GDXray \cite{mery2015gdxray}, SIXray \cite{miao2019sixray}).

Commonly, it is challenging to collect sufficient X-ray imagery containing example of prohibited items with large variations in pose, scale and item construction. To overcome this challenge, contemporary data augmentation schemes such as image translation, rotation, flipping and re-scaling are applied to enlarge the availability of otherwise limited training datasets \cite{krizhevsky2012imagenet}. However, such methods suffer from the fact that the resulting augmented dataset still lacks diversity in terms of prohibited item variation and inter-occlusion emplacement within complex and cluttered X-ray security imagery. This motivates the use of synthetically composed imagery, where such imagery readily enables the introduction of more variability in  pose, scale and prohibited item usage in an efficient and readily available way.

In this work, we devise a Synthetically Composited (SC) data augmentation approach via the use of Threat Image Projection (TIP). TIP is an established process within operational aviation security for the monitoring of human operators which uses a smaller collection of X-ray imagery comprising of isolated prohibited objects (only), which are subsequently superimposed onto more readily available benign X-ray security imagery. Here this approach additionally facilitates the generation of synthetic, yet realistic prohibited X-ray security imagery for the purpose of CNN training. Our key contributions are the following: (a) the synthesis of high quality prohibited images from benign X-ray imagery using a documented TIP approach and (b) an extended comparative evaluation on how real and synthetically generated X-ray imagery impacts the performance for prohibited object detection and classification using CNN architectures.

%% file: 2.relatedwork.tex
\vspace{-0.4cm}
\section{Related Work} \label{sec:relwork}
\vspace{-0.2cm}
Traditional computer vision methods that rely on handcrafted features have been applied to prohibited item detection in X-ray security imagery
such as Bag of Visual Words (BoVW) \cite{Turcsany2013Improving, mery2013automated, kundegorski2016using} and sparse representations \cite{mery2015object}. However, the recent advancement in CNN have drawn more attention to prohibited item detection due to significant performance gains within X-ray security imagery \cite{akccay2016transfer, akcay2018using, mery2016modern}. The works of \cite{mery2016modern, akccay2016transfer} compare handcrafted features with a BoVW based sparse representation to CNN features. These shows that such deep CNN features achieve superior performance with more than 95\% accuracy for prohibited item detection. The study of \cite{akccay2016transfer} exhaustively compares various CNN architectures to evaluate the impact of network complexity on overall performance. Fine tuning the entire network architecture for this problem domain yields 0.99\% true positive, 0.01\% false positive and 0.994\% accuracy for generalized prohibited item detection \cite{akccay2016transfer}.

Further work on prohibited item under X-ray security imagery is undertaken by Mery \etal \cite{mery2013automated}, where regions of interest detection is performed across multiple views of the object. Subsequently, the candidate region obtained from an earlier segmentation step is then matched based on their similarity. This achieves 94.3\% true positive and 5.6\% false positive across multiple view X-ray security imagery. The work of \cite{akcay2018using} examines the relative performance of traditional sliding window driven CNN detection model based on \cite{akccay2016transfer} against contemporary region-based and single forward-pass based CNN variants such as Faster R-CNN \cite{ren2015faster}, R-FCN \cite{dai2016r}, and YOLOv2 \cite{redmon2017yolo9000},  achieving a maximal 0.88 and 0.97 mAP over 6-class object detection and 2-class firearm detection problems respectively.

To investigate the generalised applicability of CNN within X-ray security imagery, large X-ray imagery datasets are required. Existing public domain  datasets such as GDXray \cite{mery2015gdxray} contains three major categories of prohibited items, \{{\it Guns, Shurikens, Razor blades}\}. However, images in GDXray are provided with lesser clutter and overlap making object detection less challenging than in typical operational conditions. By contrast, the SIXray dataset  \cite{miao2019sixray} contains six classes, \{{\it Guns, Knives, Wrenches, Pliers, Scissors, Hammers}\}, from cluttered operational imagery. This provides more inter-occluding imagery examples but at the same time provides significantly fewer prohibited item than benign samples akin to an operational (real-world) scenario, where the presence of prohibited items is low within stream-of-commerce (largely benign) X-ray security imagery. 

To overcome the limited dataset availability, data augmentation has been used to increase overall dataset diversity. Whilst simple image data augmentation strategies such as translations, flipping and scaling do increase geometric diversity of the imagery they do not increase the appearance or content diversity of the dataset itself \cite{frid2018synthetic}. The work of \cite{yang2019data} alternatively attempts data augmentation based on an Generative Adversarial Network (GAN) approach but generates synthetic prohibited items in isolation rather than within a full cluttered X-ray security image. By contrast, the work of \cite{jain2019evaluation} utilises an approach, similar to the concept of TIP, whereby a prohibited item is superimposed into X-ray security imagery.
Therefore, in this work, we 
explore the feasibility of TIP as a data augmentation strategy to support performance enhancement and evaluation of contemporary deep CNN architectures within the context of prohibited item detection in security X-ray imagery.

%% file: 3.proposal.tex
\section{Proposed Approach} \label{sec:approach}

We investigate the use of a full TIP pipeline, based on prior work in the field \cite{neiderman2005threat,schwaninger2005aviation,schwaninger2007statistical}, to generate a range of appearance and contents based dataset variation (Section \ref{subs:TIP approach}). 
Subsequently, CNN object detection architecture is used to evaluate the TIP based data augmentation approach and compare the performance with real X-ray security imagery (Section \ref{subs:detection_Strategy})


\subsection{Synthetic X-ray Security Imagery via TIP} \label{subs:TIP approach}
\begin{figure}[h]
\centering
\includegraphics[width=0.9\linewidth]{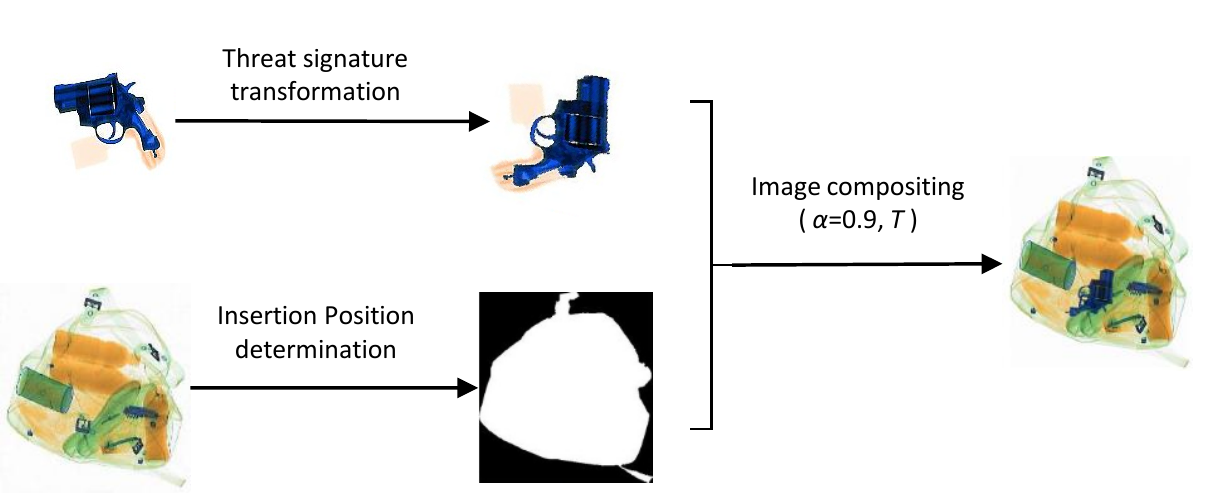}
\caption{Threat image projection (TIP) pipeline for synthetically composited image generation.}
\label{Fig:TIP_diagram}
\end{figure}

Our TIP pipeline consists of three components: \textit{threat image transformation}, \textit{insertion position determination} and \textit{image compositing} as illustrated in Figure \ref{Fig:TIP_diagram}.

\begin{figure}[t]
\centering
\includegraphics[width=\linewidth]{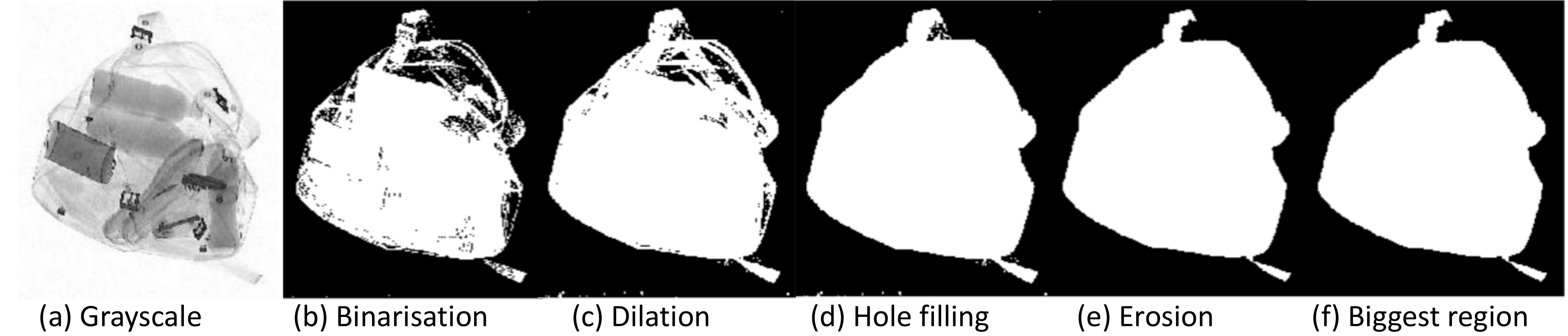}
\caption{Image segmentation using morphological operations for insertion position determination.}
\label{Fig:bagSeg}
\end{figure}

We use threat (prohibited item) images containing clean, isolated object signatures which can be easily segmented from their plain background via simple thresholding. To diversify the resultant synthetic images, we apply \textit{threat image transformation} via rotating the threat signature by a random angle $\theta$. Although other threat image transformation strategies (e.g., noise, illumination, magnification, etc.) have been explored in \cite{rogers2016threat}, our work focuses on the pure combination of our segmented threat signature and a benign X-ray security image, isolating the effects of other data augmentation techniques. We denote this transformed threat image as $I_{s}$ and the $i$-th row, $j$-th column pixel as $I_{s}(i,j)$.

A valid insertion position within the bag image is determined based on the bag region and the shape of threat signature. Given a bag image $I_{t}$, we use morphological operations to extract the bag region. Specifically, the original bag image is firstly binarised by thresholding (Figure \ref{Fig:bagSeg}b) to extract the foreground (target) region for insertion. Due to noise, a simple thresholding process cannot ideally separate background and foreground. We sequentially apply a series of appropriately parameterised morphological operations including dilation (Figure \ref{Fig:bagSeg}c), hole filling (Figure \ref{Fig:bagSeg}d) and erosion (Figure \ref{Fig:bagSeg}e) to identify the largest connected image region as the target for insertion (see Figure \ref{Fig:bagSeg}f). Obviously, a valid insertion of the threat signature has to guarantee the threat signature is completely located inside this target region. To this end, we use a loop to generate a random insertion position until it is a valid one. The selected valid insertion position can be denoted by a binary mask matrix $M$ of the same size as the target baggage image with elements of ones indicating the insertion region.

Finally, a threat signature $I_{s}$ is superimposed onto the target bag image $I_t$ in the selected valid position (denoted by $M$) to generate a synthetically composited image $I_{TIP}$. To ensure the plausibility of the composited TIP image, we consider two factors in image blend. Parameter $\alpha$ controls the transparency of the source image $I_{s}$ ($\alpha=0.9$). The other parameter is the \textit{threat threshold} $T$ ensuring the consistency of source image with the target image in terms of image contrast. The use of \textit{threat threshold} $T$ aims to remove the high-value pixels of the threat signature so that the inserted threat signature is not visually too bright comparing against the target region where it is superimposed. To calculate the value of $T$, we first transform the target image $I_t$ to a greyscale image $G_t$. The threat threshold $T$ can be empirically calculated by:

\begin{equation}
     \label{eq:threatThreshold}
     T = min(\exp{(\hat{g}^5)} - 0.5, 0.95)
 \end{equation}
 where $\hat{g}$ is the normalised average intensity of the insertion region within $G_t$ calculated as: 
 \begin{equation}
     \label{eq:avgIntensity}
     \hat{g} = \frac{\sum_{i,j} G_{t}(i,j)*M(i,j)}{\sum_{i,j}255*M(i,j)} \in [0,1]
 \end{equation}

The image compositing can be formulated as follows:
 \begin{equation}
     \label{eq:compositing}
     I_{TIP}(i,j) = \left \{
\begin{array}{ll}
     (1-\alpha) I_{t}(i,j) + \alpha I_{s}(i',j'), & M(i,j)=1 \quad and \quad I_{s}(i',j')<T*255,\\
I_{t}(i,j), & otherwise
\end{array}
\right.
 \end{equation}
 where $I(i,j)$ denotes the value of pixel in $i$-th row and $j$-th column of the image $I$; $I_s(i',j')$ denotes the pixel in source image corresponding to the pixel of $I(i,j)$. Since the value of $T$ computed by Eq.(\ref{eq:threatThreshold}) is in the range of 0.5-0.95, any pixel of the higher value than $T*255$ in the source image will be ignored during image compositing process.
 
 The proposed TIP approach is able to generate a large number of diverse synthetic X-ray baggage images containing prohibited items whose locations are accessible without any extra cost for training a supervised learning detection model.
\subsection{Detection Strategies} 
\label{subs:detection_Strategy}
We use two representative CNN object detection model, Faster R-CNN \cite{ren2015faster} and RetinaNet \cite{lin2017focal}, for the purposes of our evaluation. \\
{\bf Faster R-CNN} \cite{ren2015faster}
is an object detection algorithm which is the combination of its predecessor
Fast R-CNN \cite{girshick2015fast} and Region Proposal Network (RPN). Unlike Fast R-CNN \cite{girshick2015fast}, which utilises external region proposal, this architecture has its own region proposal network, which is consists of convolutional layers 
that generate object proposals and two fully connected layers that predict coordinates of bounding boxes. The corresponding locations and bounding boxes are then fed into objectness classification and bounding box regression layers.  Finally the objectness classification layer classify whether a given region proposal is an object or a background region while a  bounding box regression layer predicts object localisation, at the end of 
the overall detection process. \\
{\bf RetinaNet} \cite{lin2017focal} is an object detector where the key idea is to solve the extreme class imbalance between foreground and background classes. To improve the performance, RetinaNet employs a novel loss function called Focal Loss, where it modifies the cross-entropy loss such that it down-weights the loss in easy negative samples so that the loss is focusing on the sparse set of hard samples. Unlike Faster R-CNN \cite{ren2015faster} which apply two-stage approach, RetinaNet only apply one-stage approach, potentially to be faster and simpler.


%% file: 4.experimen.tex
\section{Experimental Setup} \label{sec:expsetup}
Our experimental setup comprises of real X-ray security imagery dataset and one constructed using the TIP based synthetic compositing approach outlined in Section \ref{subs:TIP approach}. These are evaluated within a common CNN training environment using the CNN architecture outlined in Section \ref{subs:detection_Strategy}. \\
\begin{wrapfigure}{r}{7cm}
\includegraphics[width=7cm]{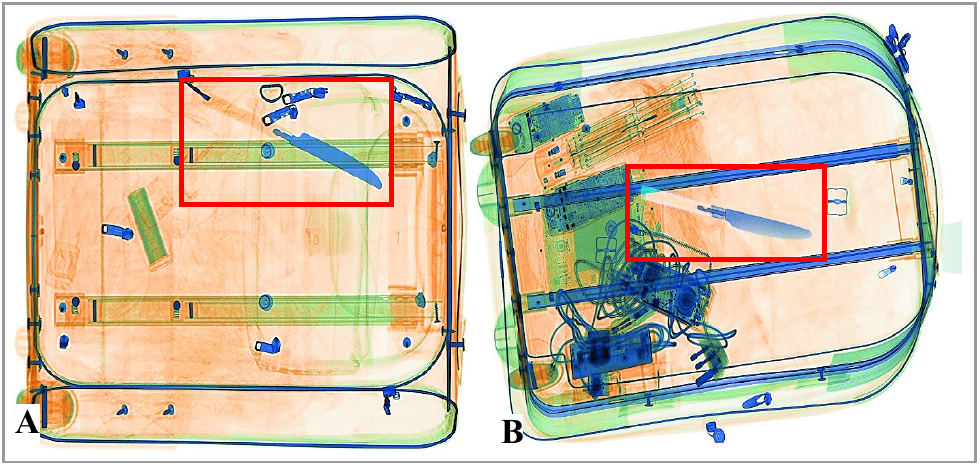}
\vspace{-0.8cm}
\caption{Visual comparison of real (A) and SC (B) X-ray security imagery of prohibited items.}
\label{Fig:threat_real}
\end{wrapfigure}
{\bf Dbf3$_{Real}$ dataset:} The Durham Dataset Full Three-class \textit{(Dbf3)} images are generated using a Smith Detection dual-energy X-ray scanner (Figure. \ref{Fig:SOC_approach}). It consists of total 7,603 images, which is divided into three classes of  prohibited item. In this experiment we uses subsets of the datasets, which consists of 
three types metallic prohibited item, \{{\it Firearm, Firearm Parts, Knives}\}. Out of these three classes, we incorporate
3,192 images of firearms, 1,204 images of firearms parts, and 3,207 images of knives, within cluttered and complex X-ray security dataset. \\
{\bf Dbf3$_{SC}$ dataset:} The Synthetically Composited (SC) dataset is generated using TIP approach of Section \ref{subs:TIP approach}. We use 3,366 benign X-ray security images, generated by a Smith Detection X-ray scanner, and 123 individual prohibited objects of three classes
\{{\it Firearm, Firearm Parts, Knives}\}. The prohibited item are composed into the benign images to create synthetically composited X-ray security imagery dataset. We use the same number of images as {\it Dbf3$_{Real}$} in the synthetically composited dataset.
Exemplar images from {\it Dbf3$_{Real}$} (Figure. \ref{Fig:threat_real}A) and {\it Dbf3$_{SC}$} (Figure \ref{Fig:threat_real}B) are visually realistic and challenging to distinguish from the real images. \\ 
{\bf Dbf3$_{Real+SC}$ dataset:} A subset of {\it Dbf3$_{Real}$} and subset of {\it Dbf3$_{SC}$} images are combined to create this dataset, where the numbers of synthetic 
and real images are are used in equal number to present a data set with 50\% of each which is itself the same size as {\it Dbf3$_{Real}$}.



The CNN architecture (Section \ref{subs:detection_Strategy}) are trained on a GTX 1080Ti GPU, optimised by Stochastic Gradient Descent (SGD) with a weight decay of 0.0001, learning rate of 0.01 and termination at maximum of 180k epochs. ResNet$_{50}$ and ResNet$_{101}$ are chosen as network backbone to operate within the detection framework of \cite{Detectron2018}.
We split each dataset into training (60\%), validation (20\%) and test sets (20\%) so that each split has similar class distribution. All CNN architecture are initialised with ImageNet \cite{krizhevsky2012imagenet} pre-trained weights  for their respective model \cite{xie2017aggregated}.

%% file: 5.evaluation.tex
\section{Evaluation} \label{sec:evaluation}

Our evaluation considers the comparative performance of CNN architecture to detect prohibited items using real X-ray imagery against prohibited items under synthetic X-ray imagery. We consider mean Average Precision (mAP) as our evaluation criteria following \cite{akcay2018using}.



\subsection{Prohibited Item Detection Results}
In the first set of experiments (Table \ref{Table:mAP_dbf3_soc}, upper), prohibited items in X-ray security imagery are detected using 
the CNN architectures set out in Section \ref{subs:detection_Strategy}.
We use the {\it Dbf3} dataset consisting of three types of prohibited items \{{\it Firearm, Firearm Parts, Knives}\}. To provide performance benchmark, our CNN architectures are firstly trained and evaluated on real images of \textit{Dbf3} ({\it Dbf3$_{Real}$ $\Rightarrow$ Dbf3$_{Real}$}). The AP/mAP highlighted in Table \ref{Table:mAP_dbf3_soc}(upper) denotes the maximal performance achieved. 
Table \ref{Table:mAP_dbf3_soc} shows statistical results of prohibited item detection for Faster R-CNN \cite{ren2015faster} and RetinaNet \cite{lin2017focal} architecture using ResNet$_{50}$ and ResNet$_{101}$. Inline with the overall complexity of the network, we observe maximal mAP performance from ResNet$_{101}$ for all three prohibited item classes. In this performance benchmark, we observe that the best performance (mAP = 0.88) is achieved on $\textit{Dbf3}_{real}$ by Faster R-CNN with ResNet$_{101}$ configuration, as presented in Table \ref{Table:mAP_dbf3_soc} (upper).

\begin{table}[ht]
\centering
\begin{tabular}{llllcll}
\cline{1-7}
\multirow{2}{*}{\shortstack[l]{Train $\Rightarrow$ \\ Evaluation}} & \multirow{2}{*}{Model} &  \multirow{2}{*}{Network} & \multicolumn{3}{c}{Average precision} & \multirow{2}{*}{mAP} \\ \cline{4-6}
& &  & Firearm & Firearm Parts & Knives &  \\  \hline
\multirow{4}{*}{\scriptsize \shortstack[l]{\it Dbf3$_{Real}$ $\Rightarrow$ \\ \it Dbf3$_{Real}$}} & \multirow{2}{*}{\small{\shortstack[l]{Faster \\ R-CNN \cite{ren2015faster}}}}  & {\small ResNet$_{50}$}  & 0.87 & 0.84 & 0.76 & 0.82 \\ 
& &  {\small ResNet$_{101}$}    & {\bf 0.91}  & {\bf 0.88}  & {\bf 0.85} & {\bf 0.88} \\ \cline{2-7}
& \multirow{2}{*}{\small RetinaNet \cite{lin2017focal}} & {\small ResNet$_{50}$} & 0.88 & 0.86 & 0.73 & 0.82 \\ 
& & {\small ResNet$_{101}$}  & 0.89 & 0.86 & 0.73  & 0.83 \\ \hline
\multirow{4}{*}{\scriptsize \shortstack[l]{\it Dbf3$_{SC}$ $\Rightarrow$ \\ \it Dbf3$_{Real}$}} & \multirow{2}{*}{\small{\shortstack[l]{Faster \\ R-CNN \cite{ren2015faster}}}}  & {\small ResNet$_{50}$}  & 0.82 & 0.77 & 0.55 & 0.71  \\ 
& &  {\small ResNet$_{101}$} & {\bf 0.86} &{\bf 0.80} &{\bf 0.66} &{\bf 0.78}  \\ \cline{2-7}

& \multirow{2}{*}{\small RetinaNet \cite{lin2017focal}} & {\small ResNet$_{50}$} &0.84 &0.77 &0.53 &0.71 
 \\ 
& & {\small ResNet$_{101}$} &0.84 &0.76 &0.54 &0.72\\ \hline

\multirow{4}{*}{\scriptsize \shortstack[l]{\it Dbf3$_{Real+SC}$ $\Rightarrow$ \\ \it Dbf3$_{Real}$}} & \multirow{2}{*}{\small{\shortstack[l]{Faster \\ R-CNN \cite{ren2015faster}}}}  & {\small ResNet$_{50}$} & 0.85 & 0.79 & 0.65 & 0.76 \\ 
& &  {\small ResNet$_{101}$} & {\bf 0.87} & {\bf 0.81} & {\bf 0.74} & {\bf 0.81} \\ \cline{2-7}

& \multirow{2}{*}{\small RetinaNet \cite{lin2017focal}} & {\small ResNet$_{50}$} & 0.85 & 0.81 & 0.64 & 0.76  \\ 
& & {\small ResNet$_{101}$} & 0.86 & 0.80 & 0.63 & 0.76 \\ \hline
\end{tabular}
\caption{Detection results of varying CNN architecture trained on: Upper $\rightarrow$ {\it Dbf3$_{Real}$}, Middle $\rightarrow$ {\it Dbf3$_{SC}$} and Lower $\rightarrow$ {\it Dbf3$_{Real+SC}$}. All models are evaluated on set of real X-ray security imagery.}
\label{Table:mAP_dbf3_soc}
\end{table}



In second set of experiments (Table \ref{Table:mAP_dbf3_soc}, middle), the CNN architecture are trained on the synthetic X-ray imagery ({\it Dbf3$_{SC}$}) achieve 0.78 mAP when tested on same set of real X-ray imagery ({\it Dbf3$_{Real}$}) of Table \ref{Table:mAP_dbf3_soc} (upper).
Even though the performance is lesser when compared with former results (Table \ref{Table:mAP_dbf3_soc}, upper), this experimental setting does not require any manual image labelling (as TIP insertion positions are known) and yet achieves surprisingly good performance on a standard benchmark. The performance gap between CNN architecture trained on real and synthetically composited X-ray  imagery is attributable to the domain shift problem whereby the distribution of training and test data differ. In the first experiment (Table \ref{Table:mAP_dbf3_soc}, upper), the training and test data are from the same distribution since they are created by randomly dividing data captured under the same experimental conditions. By contrast, in this second experimental setup (Table \ref{Table:mAP_dbf3_soc}, middle), the prohibited items used for the synthetic X-ray imagery ({\it Dbf3$_{SC}$}) data are different from those in the test X-ray imagery ({\it Dbf3$_{Real}$}) data. It is also noteworthy that prohibited images used for generating synthetic X-ray imagery ({\it Dbf3$_{SC}$}) data is a smaller set of prohibited item instances than in the real training images. As a result, CNN architecture trained on synthetic data have larger generalisation errors than those trained on real data. However, when tested on synthetic X-ray imagery ({\it Dbf3$_{SC}$}) data (Table \ref{Table:mAP_dbf3_tip}), however, CNN architecture trained with real or synthetic CNN architecture have comparable performance. These experimental results show that it is essential to have diverse prohibited item signatures in the training data to improve the generalisation. It also largely explains why overall performance in Table \ref{Table:mAP_dbf3_tip} (showing evaluation on the synthetic dataset, {\it Dbf3$_{SC}$}) is significantly higher than overall performance in Table \ref{Table:mAP_dbf3_soc} (evaluation is on the real dataset, {\it Dbf3$_{Real}$}).

\begin{table}[ht]
\centering
\begin{tabular}{llllcll}
\cline{1-7}
\multirow{2}{*}{\shortstack[l]{Train $\Rightarrow$ \\ Evaluation}} & \multirow{2}{*}{Model} &  \multirow{2}{*}{Network} & \multicolumn{3}{c}{Average precision} & \multirow{2}{*}{mAP} \\ \cline{4-6}
& &  & Firearm & Firearm Parts & Knives &  \\  \hline
\multirow{4}{*}{\scriptsize \shortstack[l]{\it Dbf3$_{Real}$ $\Rightarrow$ \\ \it Dbf3$_{SC}$}} & \multirow{2}{*}{\small{\shortstack[l]{Faster \\ R-CNN \cite{ren2015faster}}}}  & {\small ResNet$_{50}$}  & 0.88 & 0.87 & 0.84 & 0.87 \\ 
& &  {\small ResNet$_{101}$} & {\bf 0.92} & {\bf 0.92} & {\bf 0.89} & {\bf 0.91}  \\ \cline{2-7}
& \multirow{2}{*}{\small RetinaNet \cite{lin2017focal}} & {\small ResNet$_{50}$} & 0.89 & 0.87 & 0.83 & 0.86 \\ 
& & {\small ResNet$_{101}$} & 0.90 & 0.88 & 0.85 & 0.88 \\ \hline

\multirow{4}{*}{\scriptsize \shortstack[l]{\it Dbf3$_{SC}$ $\Rightarrow$ \\ \it Dbf3$_{SC}$}} & \multirow{2}{*}{\small{\shortstack[l]{Faster \\ R-CNN \cite{ren2015faster}}}}  & {\small ResNet$_{50}$}  &0.90 &0.88 &0.83 &0.87 \\ 
& &  {\small ResNet$_{101}$} & {\bf 0.93} & {\bf 0.92} & {\bf 0.86} & {\bf 0.91} \\ \cline{2-7}

& \multirow{2}{*}{\small RetinaNet \cite{lin2017focal}} & {\small ResNet$_{50}$} & 0.91 &0.89 &0.84 &0.88 \\ 
& & {\small ResNet$_{101}$} & 0.91 &0.89 &0.83 &0.86  \\ \hline 

\multirow{4}{*}{\scriptsize \shortstack[l]{\it Dbf3$_{Real+SC}$ $\Rightarrow$ \\ \it Dbf3$_{SC}$}} & \multirow{2}{*}{\small{\shortstack[l]{Faster \\ R-CNN \cite{ren2015faster}}}} & {\small ResNet$_{50}$} & 0.89 & 0. 86 & 0.83 & 0.86 \\ 
& &  {\small ResNet$_{101}$} & {\bf 0.91} & {\bf 0.89} & {\bf 0.87} & {\bf 0.89}  \\ \cline{2-7}
& \multirow{2}{*}{\small RetinaNet \cite{lin2017focal}} & {\small ResNet$_{50}$} & 0.90 & 0.87 & 0.83 & 0.87 \\ 
& & {\small ResNet$_{101}$} & 0.90 & 0.88 & 0.84 & 0.87 \\ \hline
\end{tabular}
\caption{Detection results of different CNN architecture trained on: Upper $\rightarrow$ {\it Dbf3$_{Real}$}, Middle $\rightarrow$ {\it Dbf3$_{SC}$} and Lower $\rightarrow$ {\it Dbf3$_{Real+SC}$}. All models are evaluated on set of SC dataset.}
\label{Table:mAP_dbf3_tip}
\end{table}

In the third set of experiments (Table \ref{Table:mAP_dbf3_soc}, bottom), we evaluate the effectiveness of synthetic X-ray imagery by combining it with real images of \textit{Dbf3} to create \textit{Dbf3$_{Real+SC}$} dataset, as explained in the Section \ref{sec:expsetup}. We evaluate the testing sets of images from real \textit{Dbf3} (Table \ref{Table:mAP_dbf3_soc}) and synthetically composite (Table \ref{Table:mAP_dbf3_tip}) datasets. Surprisingly, the combination of real and synthetic imagery data does not improve the results (e.g. 0.81 vs 0.88 on \textit{$Dbf3_{Real}$} and 0.89 vs 0.91 on \textit{$Dbf3_{SC}$} with Faster R-CNN and ResNet101). This can also be explained by the domain shift problem mentioned previously. Possibly this data combination can perform well if we apply domain adaptation techniques \cite{wang2019unifying} explicitly. In addition, we may also need to evaluate the quality of the TIP solution that underpins our work further.



\subsection{Qualitative Examples} \label{ssec:det_emp}
\begin{figure}[h]
\centering
\includegraphics[width=\linewidth]{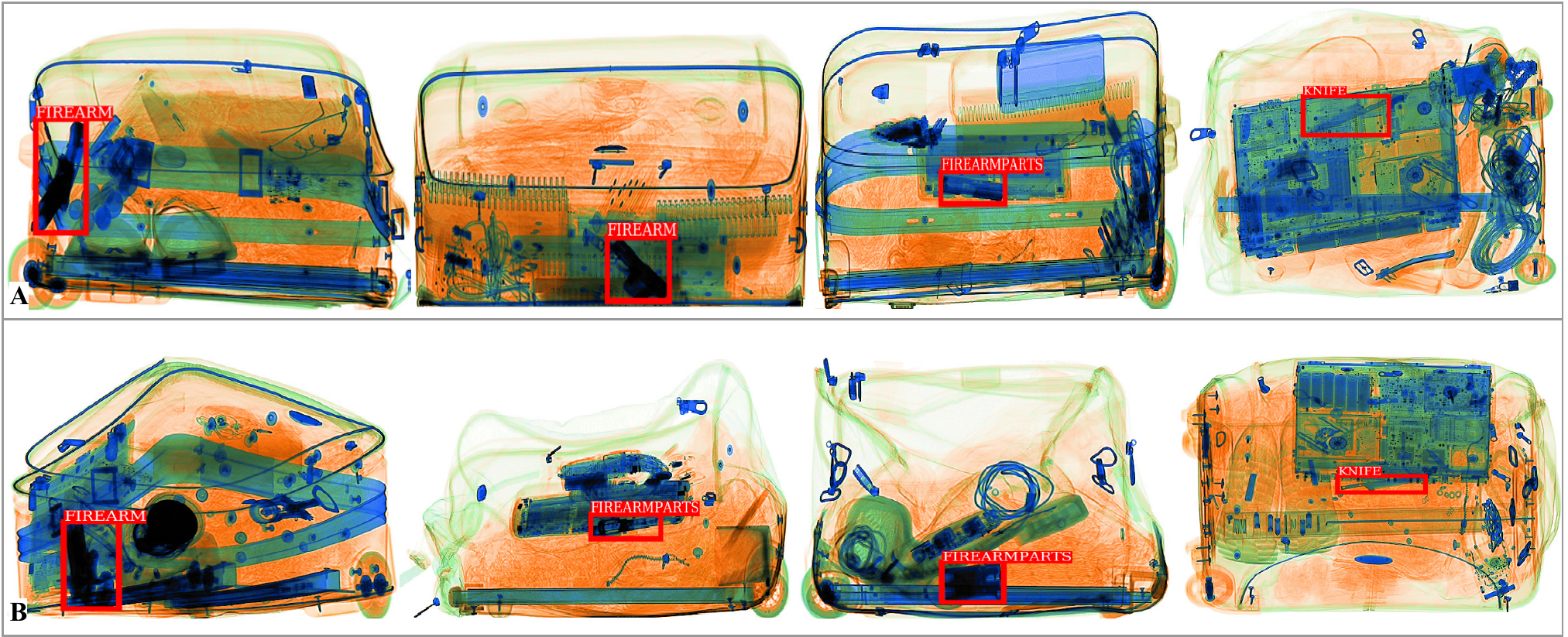}
\caption{Exemplar detection of prohibited items in red box using Faster R-CNN \cite{ren2015faster} and trained on (A) {\it Dbf3$_{Real}$} and (B) {\it Dbf3$_{SC}$} images.}
\label{Fig:obj_det_ex}
\end{figure}

Exemplar prohibited items detection results from Faster R-CNN \cite{ren2015faster} with ResNet$_{101}$ are depicted in Figure \ref{Fig:obj_det_ex}, using real (top row) and synthetic (bottom row) training imagery. 
These results illustrate that the synthetically composited imagery using TIP techniques can be effective in training detection architectures for prohibited item detection in cluttered X-ray security imagery.

\begin{figure}[h]
\centering
\includegraphics[width=\linewidth]{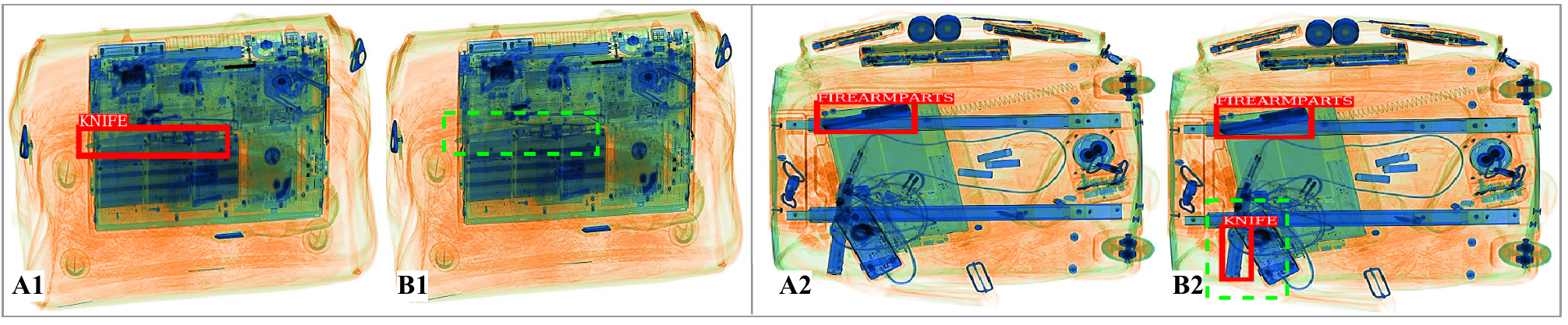}
\caption{Exemplar prohibited item detection (by Faster R-CNN \cite{ren2015faster}) using {\it Dbf3$_{Real}$} (A1,A2) and {\it Dbf3$_{SC}$} (B1,B2) training datasets. Green dashed box in B1 fails to detect, where in B2 wrongly detects as {\it knife}.}
\label{Fig:det_ex}
\end{figure}

We also visually inspect the detection results to investigate the performance difference when training the models using real and synthetic data. By comparing the results depicted in Figures \ref{Fig:det_ex}A1 and \ref{Fig:det_ex}B1, the model trained with synthetic data fails to detect the knives since such type of knives have very different appearance from the ones we used to generate the synthetic imagery. On the other hand, from Figures \ref{Fig:det_ex}A2 and \ref{Fig:det_ex}B2 we can see that the model trained on synthetic imagery has mistakenly detected something benign as a knife. These results account for the low performance for knife detection observed in Table \ref{Table:mAP_dbf3_soc}. As a result, we need to either use more diverse threat signatures for data synthesis or particular domain adaptation techniques to tackle the potential domain shift problem identified previously.

%% file: 6.conclusion.tex
\section{Conclusion} \label{sec:conclusion}
This work explores the possibility of generating synthetically composite X-ray security imagery for training of CNN architecture to bypass the collecting a large amount of hand-annotated real-world X-ray baggage imagery. 
We synthesise high-quality synthetically composited X-ray images using TIP approach and we present an extensive comparison on how real and synthetic X-ray security imagery affects the performance of CNN architecture for prohibited object detection in cluttered X-ray baggage images. Our experimental comparison demonstrates Faster R-CNN achieves the highest performance with mAP: 0.88 when trained on {\it Real} data (the good), followed by {\it Real+Synthetic} (the bad) and {\it Synthetic} (the ugly) over a three-class, \{{\it Firearms, Firearm parts, Knives}\}, prohibited item detection problem. This demonstrates a strong insight into the benefits of using real X-ray training data, also challenge and promise of using synthetic X-ray imagery.

In our future work, it is worth further investigating how to improve the effectiveness of synthetically composited imagery for training CNN architecture. Based on other work \cite{yang2019data}, a potential direction is to generate more diverse prohibited items images using generative adversarial networks (GAN). The generated prohibited item  images then could be used for generating synthetic baggage images using TIP or similar.
\\
\\
\noindent{\bf Acknowledgements}: The authors would like to thank the UK Home Office for partially funding this work. Views contained within this paper
are not necessarily those of the UK Home Office.

%% file: 7.reference.tex
{\bibliography{ref/bmvc_ref}}